\pdfoutput=1
\documentclass[11pt]{article}

\usepackage[final]{acl}

\usepackage{times}
\usepackage{latexsym}
\usepackage{multirow}
\usepackage{todonotes}
\usepackage{subcaption} % for subfigures
\usepackage{xcolor}
\usepackage{soul}
\newcommand{\highlightyellow}[1]{\sethlcolor{yellow}\hl{#1}}
\newcommand{\highlightcyan}[1]{\sethlcolor{cyan}\hl{#1}}
\newcommand{\highlightpink}[1]{\sethlcolor{pink}\hl{#1}}

\usepackage[T1]{fontenc}
\usepackage[utf8]{inputenc}

\usepackage{microtype}

\title{LePaRD: A Large-Scale Dataset of Judicial Citations to Precedent}

\author{Robert Mahari \\
 MIT and Harvard Law School \\
  \texttt{rmahari@mit.edu} \\\And
  Dominik Stammbach \\
  ETH Zurich \\
  \texttt{dominsta@ethz.ch} \\ \AND
  Elliott Ash \\
  ETH Zurich \\
  \texttt{ashe@ethz.ch} \\ \And
  Alex `Sandy' Pentland \\
   MIT \\
\texttt{pentland@mit.edu}
 }

\begin{document}
\maketitle
\begin{abstract}
We present the Legal Passage Retrieval Dataset, LePaRD. 
LePaRD contains millions of examples of U.S. federal judges citing precedent in context.
The dataset aims to facilitate work on legal passage retrieval, a challenging practice-oriented legal retrieval and reasoning task. 
Legal passage retrieval seeks to predict relevant passages from precedential court decisions given the context of a legal argument.
We extensively evaluate various approaches on LePaRD, and find that classification-based retrieval appears to work best.
Our best models only achieve a recall of 38\% when trained on data corresponding to the 10,000 most-cited passages, underscoring the difficulty of legal passage retrieval.
By publishing LePaRD, we provide a large-scale and high quality resource to foster further research on legal passage retrieval. 
We hope that research on this practice-oriented NLP task will help expand access to justice by reducing the burden associated with legal research via computational assistance.
\textit{Warning: Extracts from judicial opinions may contain offensive language.}

\hspace{3mm}
\begin{minipage}[c]{5mm}
\includegraphics[width=\linewidth]{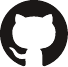}
\end{minipage}
\hspace{1.5mm}
\begin{minipage}[c]{0.7\textwidth}
\fontsize{0.76em}{0.76em}\selectfont
% Code: 
\begin{minipage}[c]{0.7\textwidth}
    \texttt{https://github.com/rmahari/LePaRD}
\end{minipage}
\end{minipage}
\vspace{-3mm}
\end{abstract}

\section{Introduction}

\begin{figure*}[ht]
\centering
\begin{tabular}{p{\textwidth}}
\multicolumn{1}{c}{\textbf{\textit{Diamond v. Chakrabarty}, 447 U.S. 303, 315 (1980)}} \\
\hline
\highlightyellow{It is, of course, correct that Congress, not the courts, must define the limits of patentability;} \highlightpink{``the province and duty of the judicial department to say what the law is.''} \highlightcyan{Marbury v. Madison, 1 Cranch 137, 177 (1803).} \\
\hline
\end{tabular}
\caption{A simple example of how judges use quotations to precedent taken from the \textit{Diamond v. Chakrabarty}. In LePaRD, \highlightyellow{preceding context} is extracted ahead of a \highlightpink{quotation} from the destination opinion (\textit{Diamond v. Chakrabarty}). Quotations are matched to the corresponding target passage from the source opinion (\textit{Marbury v. Madison}) by using the \highlightcyan{citations} contained in judicial opinions. 
The goal of legal passage retrieval is to predict the correct target passage given the preceding context.}
\label{fig:common_law_example}
\end{figure*}

A third of the global population lives in a common law jurisdiction where legal arguments are based on prior decisions, known as \emph{precedent}~\cite{FathallyMariani2008}. 
Judges and lawyers use citations to build on precedent and frequently quote passages directly from prior cases.
The U.S. legal system is an example of a common law system and U.S. federal courts have produced around 1.7 million published judicial opinions, giving rise to tens of millions of passages containing legal rules, standards, and explanations, which could potentially be cited in new cases.

Lawyers and judges frequently cite such passages as the basis for their arguments (a prototypical example is shown in Figure \ref{fig:common_law_example}).
As a result, identifying appropriate precedent relevant to a given argument represents a fundamental component of legal practice.
This is a complicated and time consuming endeavour: Based on the Case Law Access Project, a public repository of U.S. case law, there are almost 2 million published federal judicial opinions with an average length of around 4,500 tokens.
The sheer volume of passages which could potentially be cited thus adds to the complexity of legal research, which is exacerbated by subtle rules about the contexts in which a given passage is legally binding.
We provide the large-scale dataset, LePaRD, which can be used to develop computational retrieval methods that facilitate the retrieval of U.S. federal court precedent. 
LePaRD was constructed by focusing on how judges actually used precedential passages and as such it builds on millions of expert decisions.

In practice, highly paid attorneys spend significant time on legal research to find relevant precedent---and they routinely bill up to \$100 per individual search~\cite{WestlawLexisPricing2023}.
Meanwhile, in the U.S., around 90\%  of civil legal problems encountered by low-income individuals do not receive adequate legal help~\cite{LegalServicesCorporation2022} and access to such services is also limited for small businesses~\cite{baxter2022dereliction}. 
Thus, the complexity and cost of legal research may be partially responsible for the high cost of litigation and the associated access to justice gap.

Legal NLP promises to be a powerful equalizer in the legal profession~\cite{EMNLP_position_paper}, but many areas of legal practice have been slow to adopt technologies that increase efficiencies and reduce costs for clients.
While this may be partially driven by a lack of incentives and risk-aversion from legal community, legal NLP research also appears to be disconnected from the needs of legal practitioners~\cite{EMNLP_position_paper}. 
This in turn is partially driven by the lack of large-scale resources for practice-oriented legal NLP tasks.
Often, the data needed for this type of research is proprietary and constructing legal datasets from publicly available sources requires legal expertise.

To help address the high costs of legal research, and the resulting access to justice issues, and to foster more legal NLP research on practice-oriented tasks, we release the Legal Passage Retrieval Dataset LePaRD. 
LePaRD represents a large set of previously cited U.S. federal precedent, containing millions of argument contexts and the relevant target passage. 
In this work, we document the construction of LePaRD and describe relevant dataset statistics.
We also extensively evaluate various retrieval approaches from the NLP literature \citep[see e.g.,][]{anserini, reimers-2019-sentence-bert, mahari2021autolaw, tay2022transformer}, some of which have been applied to other legal information retrieval tasks \citep[e.g.,][]{Ma2021RetrievingLC, rosa2021yes}.
Our most accurate method achieves a recall of 38\% on the LePaRD test set, indicating that legal passage retrieval is a challenging task that requires new technical approaches.
No large-scale resources for legal passage retrieval exists and we address this gap by constructing and releasing LePaRD.

LePaRD contains citations to relevant precedent paired with the contexts in which they have been cited by judges. 
We also provide relevant meta-data, such as the court and decision year of an opinion, which may be relevant for future work on legal retrieval.
Retrieving relevant passages with computational assistance has the potential to reduce the time and cost associated with legal research and thus to reduce the overall cost of litigation. 
In publishing the dataset, we seek to catalyze practice-oriented legal NLP, and ultimately, and we hope that models trained on LePaRD will reduce the burden associated with legal research for litigants, judges, and lawyers, thus helping to expand access to justice. 

\section{Related Work}

Retrieval of relevant legal passages or cases is a fundamental task in legal practice. 
Most existing search tools are closed-source and the usage of such tools can cost up to \$100 per search \citep{WestlawLexisPricing2023}. 

Legal retrieval has been explored in some prior work. \citet{mahari2021autolaw} introduces the legal passage retrieval task, however, no corresponding dataset was released and the paper focused on just 5,000 target passages (in contrast to 1.8 million in LePaRD). 
This is a general problem in legal NLP where large-scale professionally annotated data sources remain proprietary\footnote{For example LexisNexis, Westlaw, and Bloomberg Law.}. 
Moreover, creating such resources remains costly due to the intricacies of legal language, which complicate the creation of large-scale resources without expert annotators who tend to be very costly. 
The lack of data has in turn made it challenging for legal NLP research to focus on tasks aligned with the needs of legal practitioners.
%Here, we introduce and release a large-scale dataset for legal passage retrieval. We find that the task becomes increasingly difficult as more target passages are included in the training set. Furthermore, we evaluate widely known and adopted retrieval algorithms from the NLP literature on LePaRD. These results allow us to situate the dataset in the literature and discuss challenges present in this task.

% coliee
Other related work includes the COLIEE shared task series related to legal case retrieval \citep[e.g.,][]{coliee_2021_summary, coliee_2022_summary}. 
In this setting, a system is given a query and has to retrieve the most related case (or statute) from a pre-defined knowledge base. 
Compared to these information retrieval tasks using synthetic queries, our dataset construction is more closely aligned with actual legal practice. 
Furthermore, the COLIEE datasets remain limited in size, containing around 4,400 cases which could potentially be retrieved\footnote{We acknowledge and greatly appreciate the continued effort in constructing and expanding the COLIEE datasets. They are increasing in size each year, however, we believe there is room for other, complementary larger-scale legal retrieval datasets.}, whereas our dataset allows us to investigate legal passage retrieval methods at scale, containing the universe of all cited legal passages in U.S. federal courts. 
This setting more closely resembles how a practicing attorney would perform legal research. 
Finally, lexical overlap seems to play a significant role in COLIEE datasets \cite{rosa2021yes}, making BM25 a strong baseline in that setting. 
In contrast, we find that this is not the case for LePaRD.

A growing body of work investigates legal citation prediction~\cite{Dadgostari2021, huang2021context} or the retrieval of relevant cases given a query~\cite{SANSONE2022101967, lecard, joshi-etal-2023-u}. 
Based on the preceding context from a legal document, the goal in legal citation prediction is to identify the citation that supports the context in question. 
By contrast, in legal passage retrieval, the aim is to identify a specific \emph{passage} of precedent rather than a citation to a whole case (which is usually tens or even hundreds of pages long).
We believe there are several reasons to focus on legal passage retrieval over legal citation prediction. 
Legal citation prediction accuracy numbers seem very strong \citep[see e.g.,][]{huang2021context}. 
We attribute these results to the long-tailed distribution of citations and believe that models take shortcuts to determine a topic for a snippet and then return the most cited cases for these topics---whereas legal passage retrieval inherently requires more involved legal reasoning. 
This also connects to \textit{relevance} in legal search, i.e., finding the appropriate target  \cite{van2017concept}. 
We believe legal \textit{relevance} is more strongly captured by searching for short passages, rather than predicting citations to entire cases, because a case is likely to deal with multiple independent arguments.

Some passages may not be semantically linked to the concepts they stand for, making it difficult to identify them using lexical overlap or semantic search.\footnote{For example the phrase ``play in the joints'' is commonly used by courts to refer to a category of state actions that are permitted by the Establishment Clause but not required by the Free Exercises Clause of the First Amendment to the U.S. Constitution, see \textit{Locke v. Davey}, 540 U.S. 712 (2004).} Instead, the link is established via frequent citations.
By contrast, sometimes there exists an entailment relation \citep[see e.g.][]{textual_entailment, snli:emnlp2015} between the context and the cited source passage, where the two passages are connected via legal reasoning. 
However, we find that this entailment in legal reasoning manifests differently in practical legal settings than in other NLP contexts. 
Thus, models trained on e.g, natural language inference \cite{snli:emnlp2015, reimers-2019-sentence-bert} fail to recognize such relations in LePaRD. 
Hence, our specially curated dataset may better facilitate the approximation of legal reasoning by NLP models.
Finally, from the perspective of practitioners, we believe that it is more useful to predict specific passages than citations to cases that may be hundreds of pages long.

\section{Legal Passage Retrieval Dataset (LePaRD)}

U.S. federal courts are bound by the doctrine of \emph{Stare Decisis}, which means that they must abide by past decisions.
As a result, judges and lawyers build their arguments on citations to precedent. 
Often these citations will be accompanied by quotations.
When performing legal research, frequently cited passages of precedent are often displayed prominently by research platforms (known as ``headnotes'' or ``key cites'') and serve as quasi-summaries of judicial opinions.
In this work, we leverage the quotations contained in judicial opinions to assemble a large dataset of precedential passages.

\subsection{Case Law Access Project} 

Harvard's Case Law Access Project (CAP) has scanned almost seven million published judicial opinions from U.S. federal and state courts.\footnote{https://case.law}
CAP provides access to raw opinion texts along with opinion metadata (which includes the relevant court, citations contained in the opinion, and the decision date).
Here we focus on judicial opinions published in U.S. federal courts including the U.S. Supreme Court, 13 federal appellate courts, and 94 district courts. 
Our study focuses on the 1.7 million published federal judicial opinions contained in CAP.

\begin{figure*}[t]
\centering
\includegraphics[width=\textwidth]{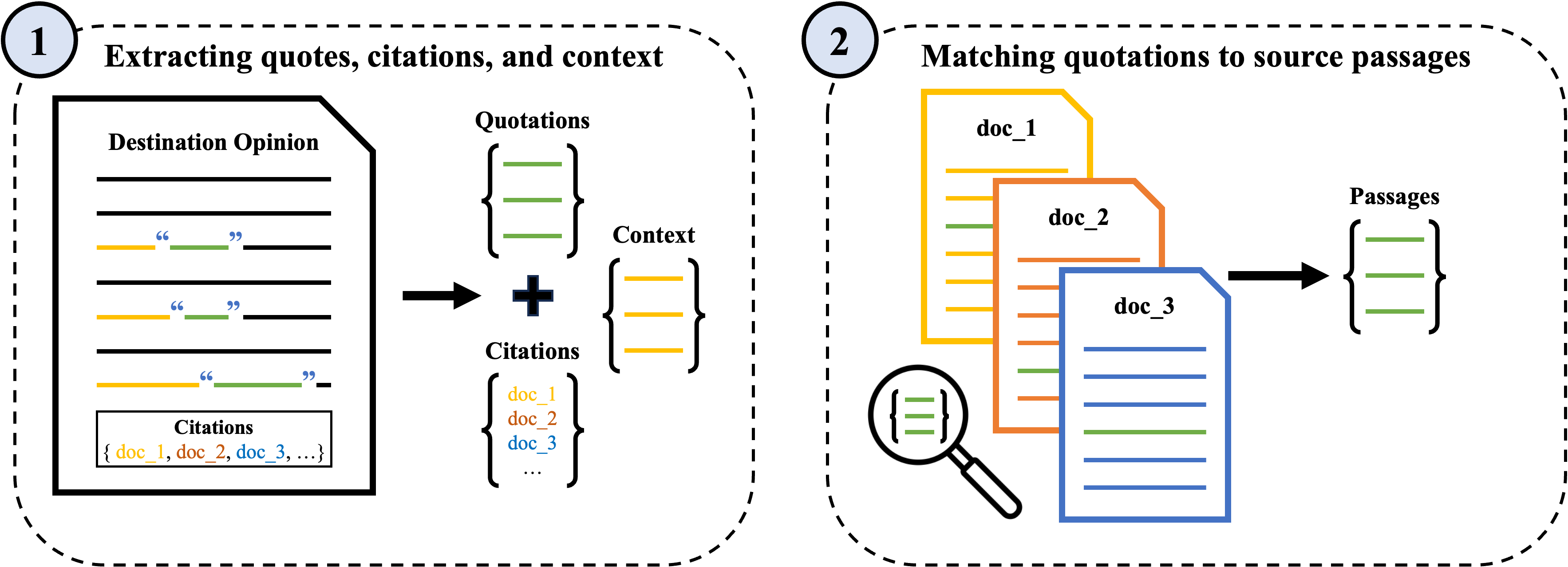}
\caption{Schematic of how LePaRD is constructed. First, we find all quotations across all 1.7 million published federal opinions in CAP and we retain the text ahead of the quotation (``context'') and the citations to other opinions. Second, we use the citations to other opinions to check whether each quotation can be matched to a passage from a prior case. If a match was found, then a training example is constructed using the relevant preceding context and the associated target passage.}
\label{fig:flowchart}
\end{figure*}

\subsection{Dataset Construction}

LePaRD is assembled by identifying quoted passages in judicial opinions, matching these passages to source opinions, and extracting the context within which the passages occur. This procedure is summarized in Figure \ref{fig:flowchart}.
In general, our construction process aims to construct a large dataset that covers as many legal contexts as possible while minimizing the amount of noise introduced by e.g., OCR errors. Given the large volume of data available, we made some design decisions that removed training examples (for example, very short passages), because including these special cases led to other issues, e.g., noisier data. 

\paragraph{Preprocessing.} For each CAP opinion, we retain the opinion id, opinion text, citations, court, and decision date. 
To facilitate downstream tasks, each opinion text was split into sentences using a Roberta model \cite{liu2019roberta} trained to predict sentence boundaries in legal text, using the legal sentence tokenization dataset introduced by \citet{sanchez2019sentence}. 
The model was trained using the transformer library \cite{wolf2020huggingfaces} with the standard hyper-parameters found in the \textit{Trainer} library. 
No further text preprocessing is performed.

For all case citations, we drop duplicated citations as well as erroneous self-citations. 
We convert citations to case ids by mapping each possible case citation to the relevant id. 
For example, \textit{Marbury v. Madison} may be cited as ``1 Cranch 137'', ``5 U.S. 137'', ``2 L. Ed. 60'', ``SCDB 1803-005'', or ``1803 U.S. LEXIS 352''. We map all of these to case\_id = 12121622.

\begin{table*}[t!]
\centering
\begin{tabular}{lrrrrrrr}
\hline
\textbf{Feature} & \textbf{Mean} & \textbf{Std} & \textbf{Min} & \textbf{25\%} & \textbf{50\%} & \textbf{75\%} & \textbf{Max} \\
\hline
Length of passage text (chars) & 275 & 192 & 22 & 167 & 233 & 326 & 13,091 \\
Length of preceding context (chars) & 719 & 647 & 5 & 171 & 454 & 1,277 & 14,275 \\
Training examples per passage & 3.62 & 33.10 & 1 & 1 & 1 & 3 & 33,439 \\
Training examples per destination & 16.20 & 30.08 & 1 & 3 & 7 & 18 & 2,576 \\
Training examples per source & 33.00 & 245.88 & 1 & 3 & 9 & 27 & 93,918 \\
Training examples per source court & 35,432 & 216,851 & 1 & 39 & 798 & 3,737 & 3,402,091 \\
\hline
\end{tabular}
\caption{Summary statistics of dataset features.}
\label{tab:dataset_summary}
\end{table*}

\paragraph{Extracting quotations and context from destination opinions.} 
For each opinions, we search for text in quotation marks (either straight or left/right quotation marks) using a regular expression.
We retain quotations longer than five words (short quotations are harder to unambiguously match to a source and may result in duplicate training data).

We extract one or more sentences of ``preceding context'' before the quotation up to a maximum of 300 words or until we reach the end of the last quotation to avoid ``overlapping contexts'' where we would have to predict multiple precedential passages from the same context.
For multi-sentence contexts, we impose this word limit as sentences vary drastically in length.

\paragraph{Matching quotations to source passages.}
We refer to the opinions from which quotations have been extracted as ``destination opinions'' and we seek to match these quotations to the relevant ``source opinion''.
Based on the previous steps, our starting point is a list of quotations and citations for each destination opinion.
Using the citations, we check whether each quotation appears in each of the cited opinions (using fuzzy string matching to account for OCR errors and modifications judges might make to the quotation to match verb tenses and capitalization).
Specifically, we match the quoted text against each sentence in the source opinion.
This means that source passages will always be a single sentence long, potentially excluding very long quotations.
In practice, we find that courts usually quote fairly short portions of longer passages (see Table \ref{tab:dataset_summary}).
To avoid many versions of the same passage we retain the entire passage sentence as the target (see Appendix \ref{sec:data_sample} for some examples.).
If a quoted passage is found to exist in a cited opinion, then this opinion is treated as the ``source'' of the passage.
Each passage thus has one source but it may have many destinations (two on average, see Table \ref{tab:dataset_summary}).
While most of the unsuccessful matches are quotations that do not come from other opinions, our approach does not tend to match multi-sentence quotations or ellipsized quotations.

LePaRD contains the preceding context, target passage, destination court, source court, destination decision date, and source decision date for each quotation that could be matched to a passage.
%The reference may be either in the form of the passage quotation or a citation to the source opinion (this is to avoid including a citation to the target passage in the context). 
Ultimately, we extract and validate 3.9 million unique target passages that have appeared in approximately 14 million contexts.

\section{Dataset Statistics}
\label{sec:data_stats}

Quotations serve several purposes in legal writing: they may be used for emphasis, refer to case documents and exhibits, introduce information from witness or expert testimony, cite supporting materials like treatises or academic publications, or they may reference precedential court opinions.
%Across all federal judicial opinions, we identify 15.9 million quotations over five words long and we successfully match 4.3 million of these to precedential passages.
While quotations and citations in judicial opinions offer several interesting avenues for legal NLP and legal passage retrieval, we focus on quotations that can be mapped to single sentences from another opinion.
Future work could also examine the retrieval of longer passages or move beyond quotations to general citations (many of which are associated with a ``pincite'' or page number).
In this section, we present several summary statistics about LePaRD (see Table \ref{tab:dataset_summary}) and we highlight some key observations.

First, we note that citations in judicial opinions obey a long-tailed distribution, with the top-1\% accounting for 24\% of all citations and 15\% of all passages receiving just 1 citation. 
This results in an inherent imbalance in the dataset, raising unique challenges for legal precedent retrieval.

Second, the sentence lengths vary substantially and this results in passages and contexts of varying lengths (the longest passage is over 13,000 characters long). 
This means that many passages and contexts will be truncated by standard text retrieval approaches. 

Third, most destination opinions contain several passages (around 16 on average, but occasionally tens or hundreds). 
This suggests that there are multiple contexts that occur within a single opinion---something that will be familiar to legal practitioners.
In our view, this validates the approach of using local context before a quotation rather than searching for more remote context that may be less relevant (for example, many opinions will discuss factors related to jurisdiction or venue early on but these will not come up anywhere else in the opinion).

Fourth, the average source opinion is represented 33 times in our data. 
While we treat passages from the same source as separate, it appears likely that they would be conceptually linked (since the portions of an opinion that are cited tend to be somewhat novel or unique and it is uncommon, though not impossible, for there to be multiple such passages in the same opinion).
Future work could thus explore whether passage retrieval benefits from grouping passages by their source. 

Finally, we find that there is a tremendous amount of variance in the training data by source court. 
We include courts to allow future users of LePaRD to narrow predictions by court in order to consider the role of binding precedent. 
However, it appears that for most courts, there is insufficient data to train independent models. 

\section{Experiments}\label{sec:experiments}

\paragraph{Problem Definition.}
Legal passage retrieval seeks to identify passages of precedent given a legal context. In total, we have around 3.9 million unique candidate passages that have been quoted at least once. Hence, given a legal context $x_{i}$, the task is to retrieve the relevant cited passage $y_{i}$ from the set of all possible passages $\{y_{1}, y_{2}, ..., y_{n}\}$.

\paragraph{Experimental Setup.}

\begin{table}[h!]
    \centering
    \footnotesize
    \begin{tabular}{l c c c}
         Number of cited passages & Train & Dev & Test \\ \hline
         10'000 & 1,732 & 95K & 95K \\ 
         20'000 & 2,228K & 124K & 124K \\
         50'000 & 3,149K & 175K & 175K \\ \hline
    \end{tabular}
    \caption{Number of examples in different splits of \mbox{LePaRD}.}
    \label{tab:dataset_statistics}
\end{table}

We release three sets of passages mapped to precedent in LePaRD. 
In these different sets, we vary the numbers of potential target passages from 10K to 50K (containing the most cited $n$ passages). 
We also release a dataset with \textit{all} cited passages to enable research on one-shot retrieval. 
This is an important extension as the majority of passages have only been cited once. 
Note that the labels from the 10K to the 20K version increase by a factor of 2, but the number of training examples only by a factor of 1.3. 
This is because citation frequency of passages obeys a long-tailed distribution where a few passages are cited with disproportionate frequency, while most are rarely cited.

We split the dataset into training, development and test sets, with 90\% of the data being in the training set, 5\% in the development set, and 5\% in the test set. We show dataset statistics in Table \ref{tab:dataset_statistics}.

We compare a variety of well-established retrieval algorithms from the NLP literature on LePaRD. 
These results are intended to serve as a baseline for follow-up work to build upon.
Our experiments highlight some of the key challenges related to legal passage retrieval and suggest that there is ample room for future work on this task.

We specifically experiment with (1) a sparse lexical retrieval approach via BM25 using the Anserini package \cite{anserini}, (2) a dense embedding-based retrieval approach using generic SBERT embeddings\footnote{We use the \href{https://huggingface.co/sentence-transformers/all-mpnet-base-v2}{all-mpnet-base-v2} model which at the time of experimenting was the best overall SBERT model across 14 benchmarks.} \cite{reimers-2019-sentence-bert}, followed by maximum dot-product similarity retrieval via the FAISS package \cite{johnson2019billion}, (3) a fine-tuned SBERT variant where we fine-tune SBERT\footnote{Using the code from the SBERT github repository with the already set hyper-parameters: \url{https://github.com/UKPLab/sentence-transformers/}.} on our training set using the Multiple Negatives Ranking Loss \cite{henderson2017efficient}, and (4) passage retrieval as a text classification task where each target passage is mapped to a unique label which is the prediction target for its preceding context \cite{mahari2021autolaw, tay2022transformer}. 
We provide results in this setting for a DistilBERT model \cite{sanh2020distilbert}, and LEGAL-BERT \cite{chalkidis-etal-2020-legal}, a domain-adapted BERT model trained on vast amounts of legal documents. The classification models have been trained using the huggingface transformer library \cite{wolf2020huggingfaces} with the standard hyper-parameters found in the \textit{Trainer} class.

The aim of our experiments is to include results for established information retrieval methods. These methods have been used extensively in all NLP domains. Our experiments are all implemented using their respective libraries and standard hyperparameters described above.

\subsection{Results}

\begin{table*}[h!]
    \small
    \centering
    \begin{tabular}{l c | c c c c | c c c c}
        \textbf{Approach} & \textbf{N} &  \multicolumn{4}{c|}{\textbf{Development Set}} & \multicolumn{4}{c}{\textbf{Test Set}} \\ 
        & & rc@1 & rc@10 & NDCG@10 & MAP & rc@1 & rc@10 & NDCG@10 & MAP \\ \hline
        \multirow{3}{*}{BM25} & 10K & 9.42 & 26.52 & 17.21 & 14.32 & 9.62 & 26.78 & 17.44 & 14.54 \\
        & 20K & 7.86 & 23.21 & 14.82 & 12.22 & 8.08 & 23.36 & 15.0 & 12.41 \\ 
        & 50K & 6.77 & 19.5 & 12.49 & 10.33 & 6.84 & 19.67 & 12.58 & 10.4 \\ \hline
        \multirow{3}{*}{SBERT} & 10K & 6.03 & 19.56 & 12.01 & 9.69 & 6.11 & 19.75 & 12.14 & 9.8 \\
        & 20K & 4.92 & 16.39 & 9.96 & 7.99 & 5.11 & 16.62 & 10.18 & 8.2 \\
        & 50K & 4.11 & 12.95 & 8.01 & 8.01 & 4.16 & 13.04 & 8.07 & 6.56 \\ \hline
        \multirow{3}{*}{fine-tuned SBERT} & 10K & 19.18 & 61.97 & 38.76 & 31.54 & 19.42 & 62.13 & 38.91 & 31.7 \\
         & 20K & 15.2 & 51.87 & 31.58 & 25.33 & 15.41 & 51.81 & 31.67 & 25.46 \\
         & 50K & 11.39 & 39.49 & 23.74 & 18.91 & 11.34 & 39.52 & 23.75 & 18.92 \\ \hline
         \multirow{3}{*}{LEGAL-BERT Classifier} & 10K & 35.45 & 79.22 & 56.86 & 49.75 & 35.68 & 79.5 & 57.08 & 49.96 \\
         & 20K & 28.62 & 68.75 & 47.74 & 41.13 & 28.45 & 68.68 & 47.63 & 41.01 \\
         & 50K & 19.32 & 48.28 & 32.93 & 28.13 & 19.24 & 48.26 & 32.92 & 28.12 \\ \hline        
        \multirow{3}{*}{DistilBERT Classifier} & 10K & 37.77 & 81.2 & 59.23 & 52.22 & 38.0   & 81.23 & 59.37 & 52.40 \\
         & 20K & 33.07 & 75.21 & 53.55 & 46.68 & 33.05 & 74.95 & 53.49 & 46.68 \\
         & 50K & 26.76 & 65.53 & 45.25 & 38.86 & 26.63 & 65.73 & 45.26 & 38.82 \\ \hline      
    \end{tabular}
    \caption{We run different approaches to the legal passage retrieval task on versions of LePaRD with a varying number of target passages (N). We measure the recall at \textit{1} and \textit{10}, normalized discounted cumulative gain (NDCG) at \textit{10}, and Mean Average Precision (MAP) for development and test sets across these baselines. The best results are obtained using classification and (relatively) few labels. Metrics are calculated using the \citet{VanGysel2018pytreceval} package.}
    \label{tab:main_results}
\end{table*}

We observe that there is only limited lexical overlap between the context and the cited passage, reflected in rather poor performance of the BM25 retrieval. This is in strong contrast to e.g., the COLIEE shared tasks where BM25 remains one of the most competitive retrieval methods \cite{rosa2021yes}. 
Deploying a pre-trained SBERT variant also seems to transfer poorly to the legal passage retrieval task. 
We attribute this finding to the domain shift (i.e., he model was not trained on legal data), and the particular challenges of legal language and entailment present in the legal passage retrieval task.

We find, however, that results improve noticeably as soon as we start to fine-tune models on the LePaRD training set. 
We see at least double the recall for dense SBERT-based retrieval after domain-specific fine-tuning. 

Recall results improve even further if we treat legal passage retrieval as a supervised classification task: Rather than seeking to embed a context and target passage close in some representation space, we assign a unique class label to each passage, and then aim to predict that label from the legal context \citep[see e.g.,][]{mahari2021autolaw, tay2022transformer}.
We experiment with two different models, and observe that the DistilBERT model achieves the best overall performance in all settings. 
Our best performance in the 10K label setting suggests that the correct target passage would be predicted among the top 10 search results in 8 out of 10 cases.

Surprisingly, a domain-specific LEGAL-BERT model achieves worse performance than the more generic DistilBERT model. We speculate that LEGAL-BERT has been pre-trained on vast amounts of legal text from various judicial systems---and some of this pre-training data does not seem to be beneficial to retrieving relevant U.S. precedent. 

Although a supervised classification approach seems to work best in our experiments, this approach comes with major limitations. 
Firstly, updating models to accommodate new precedent requires either updating existing models or re-training them from scratch~\citep{tay2022transformer}. 
Secondly, LLMs have been shown to exhibit biases \citep{abid2021persistent,lucy2021gender} and the resulting classification of passages in our application might potentially perpetuate these biases. 
Lastly, zero- and few-shot retrieval for the long tail of the distribution will not be solved by this approach, and require other methods, as highlighted by the inverse relationship between model performance and the passages frequency.

Our experiments showcase how LePaRD is a large-scale yet challenging legal retrieval dataset. 
We believe there is ample room for improvement, for example by considering re-ranking approaches or late interactions \citep{khattab2020colbert}. 
Nevertheless, our experiments help us make sense of the dataset, by e.g., highlighting how there is only limited lexical overlap between context and the target passage. 
All experiments exhibit consistent behavior across dataset splits and metrics---and are intended as baselines to be used in future research involving LePaRD.

\section{Expert Evaluation}
\label{sec:expert_eval}
A legal expert (licensed attorney) reviewed 100 randomly sampled training examples.
For each example, the expert determined whether (1) the example was generally clean and free of errors and (2) the preceding context provided sufficient information to determine that the target passage is relevant to the context.
Based on this evaluation, all examples were clean and free of errors other than preexisting errors stemming from the OCR---we leave addressing these as an opportunity for future work.
In 99\% of these examples, the expert determined that there was enough information in the context to determine the relevance of the target passage.
In the problematic case, the destination context spans two footnotes, the former a series of citations to unrelated memoranda, and the latter an explanatory footnote containing a quotation. 
Due to the CAP processing, these unrelated consecutive footnotes appear as adjacent sentences. 
Further investigation showed that this type of explanatory footnote with a quotation is very uncommon in the data.

\section{Discussion}
We highlight that the legal passage retrieval task is non-trivial, complicated by the long-tailed distribution of cited precedent and the sheer size of the corpus.
By publishing LePaRD, we aim to encourage NLP work on a set of problems that are closely aligned with the needs of the legal profession.
More broadly, our aim is to offer an example of how NLP can be used to broaden access to justice and to catalyze similar work in other legal domains.

One of the challenges of legal research is that not all case law content carries the same weight.
On the one hand, the structure of court systems means that precedent that is binding in one court may not be binding in another court, even if they are part of the same system (e.g., precedent from the U.S. District Court for the District of Massachusetts is not binding in the U.S District Court for the District of Oregon because these district courts are part of different judicial circuits within the U.S. federal judiciary). 
Similarly, old precedent may be overturned and thus lawyers must be careful to cite ``good law'' (although we find that passages tend to be cited for an average of about ten years, see Appendix \ref{sec:additional_statistics}).
On the other hand, not everything that is said in a judicial opinion has the status of \emph{precedent}: only the elements of a court's reasoning that are essential to the decision bind future courts while other content contained in a judicial opinion is known as \textit{obiter dictum} and is not legally binding.
As a result, methods that focus on lexical overlap or semantic search create a large risk of retrieving content that is not binding precedent.
LePaRD addresses these issues in two ways. 
First, we include the court and date associated with each precedent to facilitate the identification of precedent that is binding in a certain court and time.
Second, only passages that have been previously cited by judges are included in the dataset, which significantly reduces the probability of retrieving non-binding dicta. 
While we note that requiring a passage to be cited at least once restricts our dataset, we believe this limitation is far outweighed by the value of knowing that the passage has been selected for citation by a federal judge.

% RAG Discussion
One particularly promising application of precedent prediction is its potential to serve as the basis for retrieval augmented generation using large language models (RAG).
RAG has been put forward as a method of allowing models to generate text based on information that is not contained in the training data~\cite{RAG, karpukhin2020dense, FewShotRetrievalAugmented}.
In the context of legal research and writing, RAG appears to have several key advantages.
First, RAG is likely to increase the correctness of citations by allowing practitioners to ensure that only real precedent are cited (i.e., reducing, though not eliminating, the risk of hallucinations), the cited precedent is relevant to the particular court, and the cited precedent remains good law (it has not been overturned). 
The importance of this capability was highlighted by the recent \textit{Mata v. Avianca Airlines} case where an attorney relied on ChatGPT to write a brief that turned out to rely on non-existent references~\cite{Weiser2023}
Second, RAG is more easily updatable than fine-tuned models and thus allows case law to be quickly updated as new cases come out and old cases are overturned~\cite{mahari2023transparency}.
Third, RAG is auditable in the sense that practitioners see the basis for generated outputs, allowing them to remove any irrelevant precedent before text is generated.
The effective design of such systems to integrate well into lawyers' exist workflow raises interesting questions around human-computer-interaction.
While rules of professional responsibility related to lawyers' use of generative AI continue to evolve, some proposals highlight an attorney's ``duty to supervise'' the technologies they use~\cite{Greenwood2023} and the ability to evaluate what precedent will be used as a basis for a brief appears to be a likely prerequisite for ``supervising'' brief writing models. 

\section{Conclusion}

We introduce LePaRD, a large-scale dataset for predicting a target precedential passage given a legal argument context. Legal passage retrieval is an important task for legal practitioners, and a challenging NLP retrieval task. 
From a legal perspective, searching for relevant case law consumes significant resources and contributes to the cost of litigation and the associated access to justice gap.
From an NLP perspective, legal passage retrieval is a retrieval task with little lexical overlap between queries and targets, which makes it a particularly interesting retrieval problem.

We present various experiments to provide initial benchmarks and to highlight the difficulty of the legal passage retrieval task.
There are several approaches toward better legal precedent retrieval, some of which we outline here, and the experiments we present are intended as baselines rather than optimal solutions.
One example approach is to combine citation and passage retrieval to first find relevant cases and then identify specific passages within them---which can be thought of as a retrieve and re-rank approach. 
Alternatively, one could also retrieve the top-N passages, and re-rank those with a more powerful re-ranker. 
We are excited for LePaRD to serve as a large-scale resource for such experiments and other retrieval research in the legal domain.

%Other extensions:
%We publish the LPRD with the hope of encouraging work on legal passage retrieval, and in particular context-aware retrieval that takes into account the relevant court and time.
%Beyond the key application of using the LDPR
% Uncited passages
% Additional context

\section{Limitations}
We discussed several limitations of this work throughout the paper. In this section, we expand on some of these points, detail other limitations, and outline avenues for future work.
\paragraph{Noise in the CAP data.} Opinions are usually published in a PDF format. CAP converted these PDFs into text, which at times results in errors and the resulting text can contain errors typical in such conversion efforts at scale. While the data is clean enough to provide a valuable NLP dataset for retrieving relevant legal passages, and works well to explore legal information retrieval methods, it would need to be corrected for submission in a legal document.
\paragraph{Fuzzy Matching.} LePaRD is created by heuristically leveraging quotations and the case law citation to retrieve the source passage from the source opinion. Due to our heuristics, OCR errors and fuzzy matching, not all examples in the dataset are true examples a source passage being cited. In particular, if Opinion A quotes Opinion B which quotes Opinion C, then it is possible that a passage quoted in A will be matched to both B and C although it originates from C. However, after expert evaluation and several experiments, we believe that LePaRD is a high quality dataset that can form the basis of impactful NLP research.
\paragraph{Focus on the U.S. legal system.} LePaRD contains only U.S. precedent. In future work, we plan to explore whether we can create similar datasets for other jurisdictions or even for civil law contexts where citations to regulations, laws, and statues predominate. 
\paragraph{Experiments.} In the experiment section, we show experiments for, by today's standards, small transformer models such as DistilBERT and SentenceBERT. We believe that using larger and more recent models such as LLama 2~\cite{touvron2023llama} will result in better performance. However, the experiments we show are intended to be generally accessible, including for researchers with limited compute budgets. In particular, we highlight well-established retrieval algorithms, like BM25 and dense retrieval, and believe that these provide valuable baseline experiments and insights. We think of this contribution as a resource paper where we provide appropriate baseline results. Thus, we leave exploration of larger and more recent models to future work. 

\section{Ethical Considerations}
\paragraph{Intended Use.} This work presents a legal information retrieval dataset---it is not intended to be a resource for anyone engaged in a legal dispute. LePaRD is aimed to further practice-oriented legal NLP research and it also could form the basis for real-world systems that help litigants and their attorneys with legal research. We hope that the development of these types of technologies will help alleviate the access to justice crisis.

\paragraph{Misuse Potential.} We recognize that the legal context is especially sensitive, and caution researchers to think carefully about how they use LePaRD and other legal datasets. In particular, efficient legal research could help under-resourced litigants, but it can also facilitate frivolous filings. 

\paragraph{Model Bias.} Although the reported performance of NLP models is often very high, it is widely known that ML models suffer from picking up spurious correlations from data. Furthermore, it has been shown that pre-trained language models such as DistilBERT and LegalBERT suffer from inherent biases present in the pre-training data \citep{abid2021persistent, lucy2021gender}. This in turn leads to biased models---and it is thus likely that the models we present also suffer from such biases. This is especially troubling if legal passage retrieval methods work particularly poorly for certain areas of law or certain categories of litigants; we highlight the exploration of these biases and their mitigation as an important area for future work.

\paragraph{Data Privacy.}
The data used in this study is exclusively public textual data provided by CAP. It contains legal opinions from the U.S. which are public records. There is no user-related data or private data involved, which would not have been public prior to our work.

\bibliography{anthology,custom}
\bibliographystyle{acl_natbib}

\appendix
% \section{Quote lengths}

% \begin{table}[ht]
% \centering
% \begin{tabular}{c|c}
% \textbf{Percentile} & \textbf{Text Length} \\ 
% \hline
% 10 & 7 \\
% 20 & 11 \\
% 30 & 17\\
% 40 & 24\\
% 50 & 44\\
% 60 & 73 \\
% 70 & 107\\
% 80 & 149\\
% 90 & 215\\
% \hline
% \end{tabular}
% \caption{Distribution of quote lengths (number of characters)}
% \label{tab:quote_lengths}
% \end{table}

\section{Data Sample}
\label{sec:data_sample}

Table \ref{tab:LePaRD_sample} shows a sample of five training examples from LePaRD. Note how often only a small portion of a target passage is actually quoted in the destination opinions.

\section{Further Dataset Statistics}
\label{sec:additional_statistics}

Here we provide some additional insights derived from LePaRD.
In contrast to the details provided in Section \ref{sec:data_stats}, we will explore interdisciplinary insights that may catalyze future research.

\begin{figure*}[]
\centering
\includegraphics[width=\linewidth]{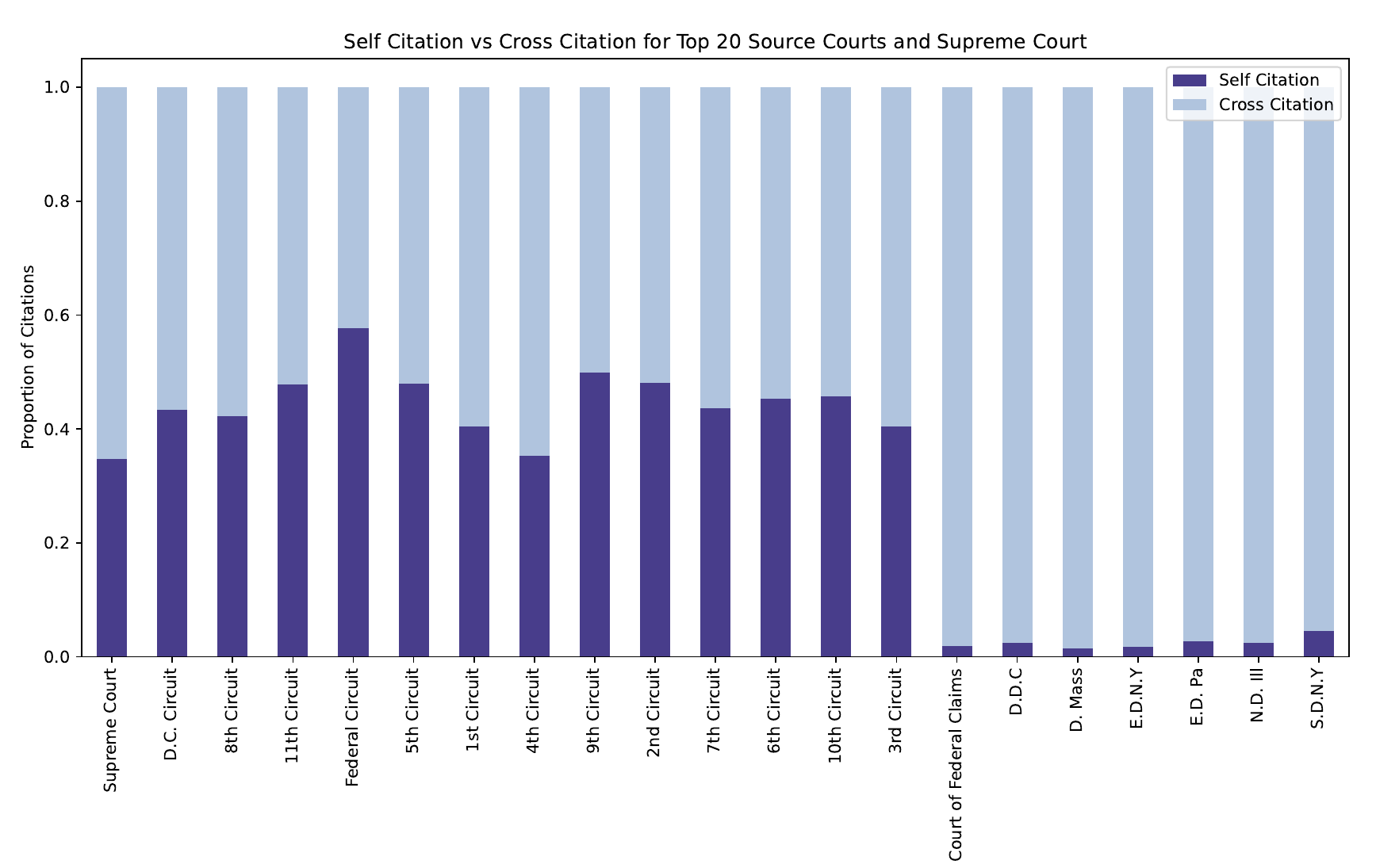}
\caption{Comparing citations to judicial opinions from the same court (``self citation'') to citations to other courts (``cross cite''). We find that appellate courts are most likely to cite themselves, while district courts only rarely cite their own precedent.}
\label{fig:cross_cite}
\end{figure*}

\begin{figure*}[]
\centering
\includegraphics[width=\linewidth]{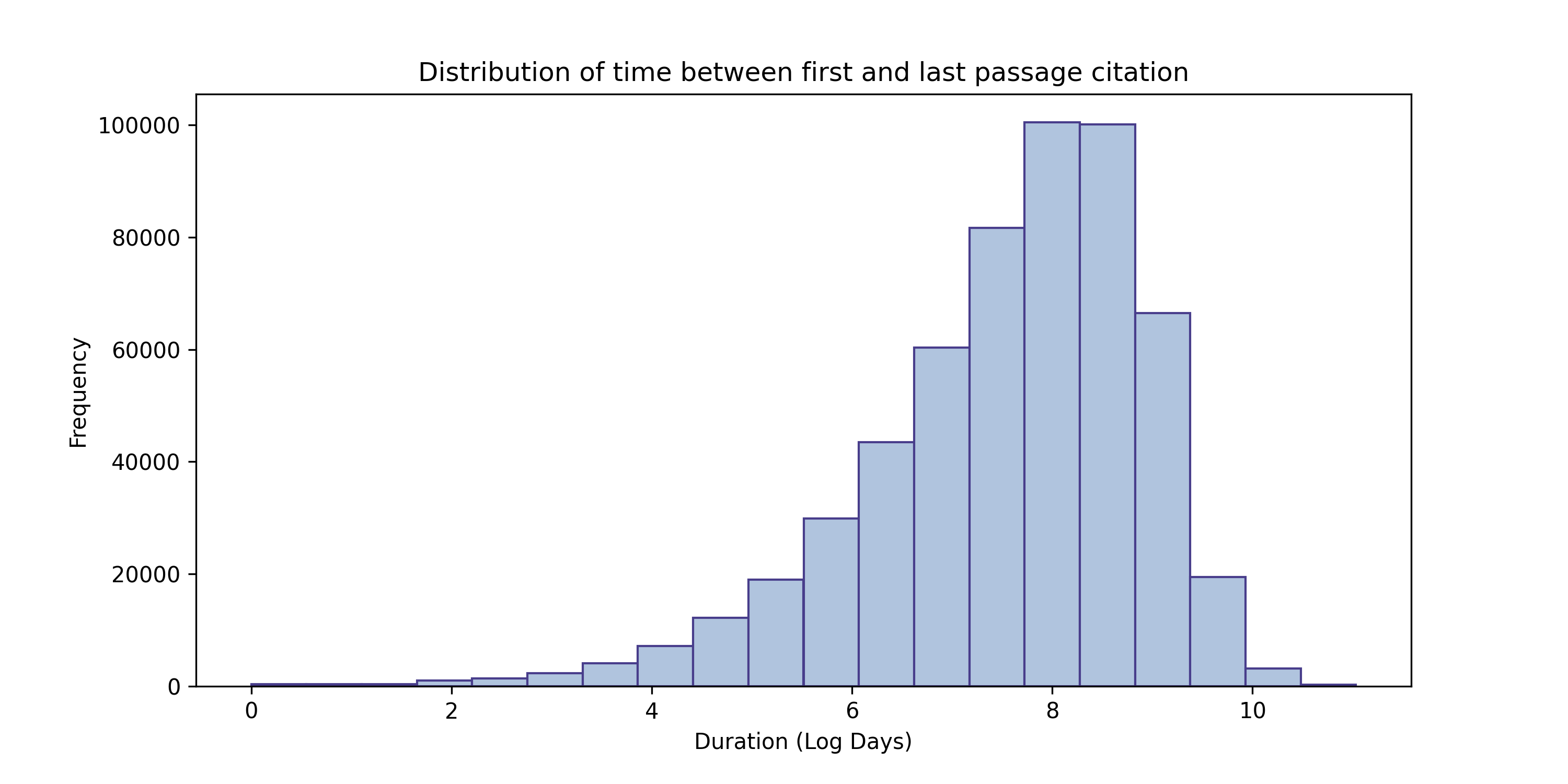}
\caption{Distribution of time in units of log days between the first and last citation of a passage in our data.}
\label{fig:duration}
\end{figure*}

We find that passages are cited for a long time after initial publication with a mean of 10 years and a maximum of over 150 years between the first and last citation (see Figure \ref{fig:duration}).
This is relevant insofar as it highlights that a legal passage dataset will be a valuable contribution with a lasting impact for legal precedent retrieval.
We further observe that a majority of quotations are to passages produced by another court, especially by the U.S. Supreme Court or by appellate courts (see Figure~\ref{fig:cross_cite}).
In particular, district courts appear to cite very little of their own precedent, which is unsurprising given that they are bound by the relevant higher courts and thus are more likely to cite precedent from these higher courts.
These observations provide some evidence that LePaRD represents a fairly representative sample of precedential passage usage in federal courts.

\begin{figure*}[t]
\centering
\includegraphics[width=\textwidth]{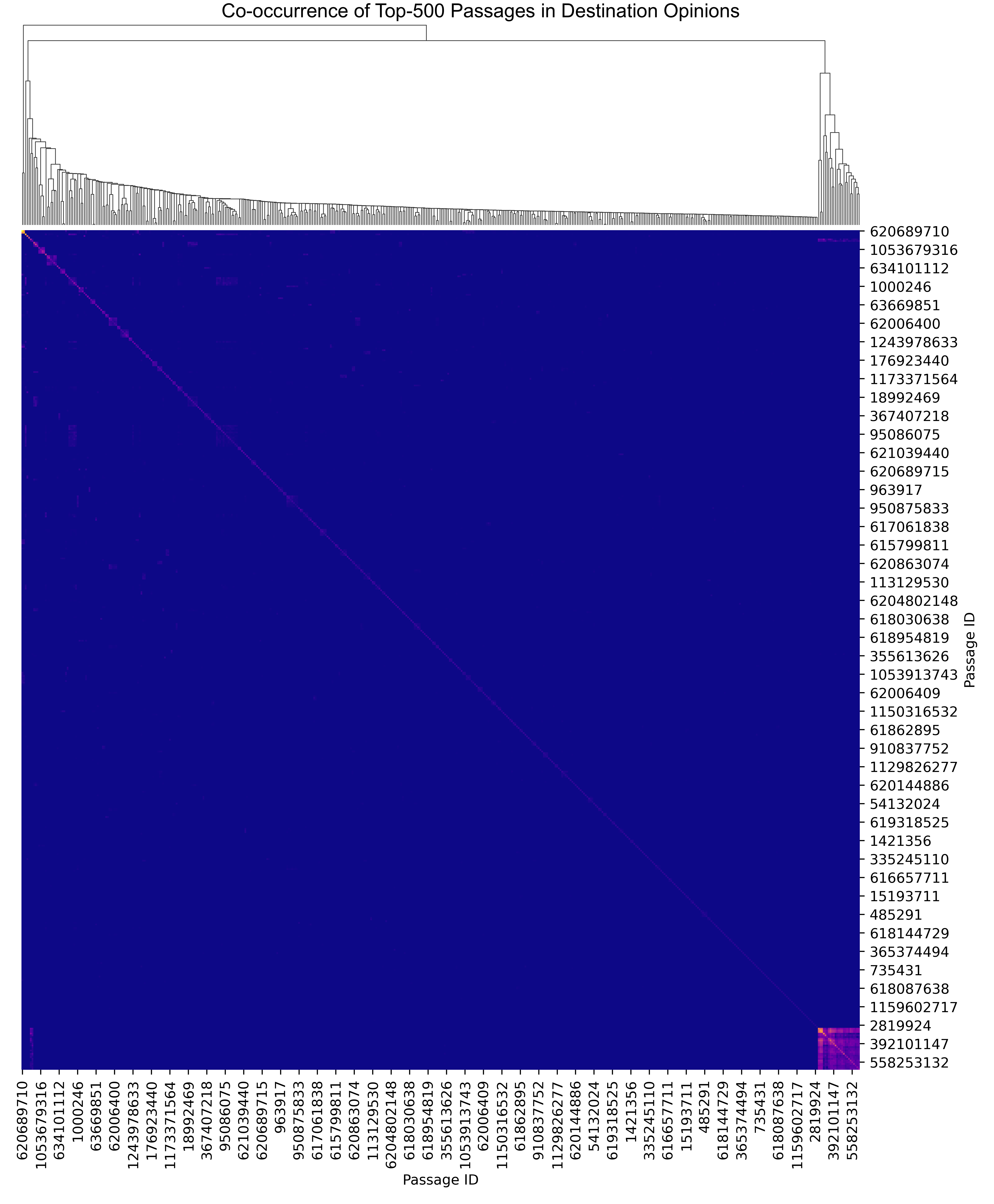}
\caption{Hierarchical clustering of passage co-occurrence.}
\label{fig:cluster}
\end{figure*}

Clustering passage co-occurrence based on whether passages appear in the same destination context reveals interesting patterns (see Figure~\ref{fig:cluster}). 
We observe three clusters: First, a very small cluster (just two cases, \textit{Anderson v. Liberty Lobby, Inc.} and \textit{Celotex Corp. v. Catrett}) which pertain to summary judgement, when a judgement is entered without a full trial which happens very frequently in many different civil disputes. 
Second, a small cluster of bankruptcy court cases, which are brought in a subset of specialized federal courts. Third, a large cluster containing all other passages.
This clustering highlights an alternative approach to legal passage retrieval that uses a pre-existing set of citations to predict missing ones, as explored by \cite{huang2021context}.

\clearpage
\newpage

\begin{table*}[]
\centering
\scriptsize
%\tiny
\begin{tabular}{|p{3.3cm}|p{4cm}|p{4cm}|}
\hline
\textbf{Meta-Data} & \textbf{Preceding Context} & \textbf{Target Passage}\\
\hline
\textbf{Destination Court:} E.D.N.Y \newline
\textbf{Destination Date:} 2001-03-28 \newline
\textbf{Source Court:} Supreme Court \newline
\textbf{Source Date:} 1974-12-23 & In order to satisfy this requirement, a plaintiff must establish a “sufficiently close nexus between the State and the challenged action. See American Mfrs. Mut. Ins. Co. v. Sullivan, 526 U.S. 40, 50, 119 S.Ct. 977, 985, 143 L.Ed.2d 130 (1999). Alternatively, if the government has & There where a private lessee, who practiced racial discrimination, leased space for a restaurant from a state parking authority in a publicly owned building, the Court held that the State had \textbf{so far insinuated itself into a position of interdependence with the restaurant that it was a joint participant in the enterprise.} \\
\hline
\textbf{Destination Court:} D.D.C. \newline
\textbf{Destination Date:} 2012-02-13 \newline
\textbf{Source Court:} Supreme Court \newline
\textbf{Source Date:} 2005-04-19 & He filed no opposition. That Order was also mailed to Plaintiff on Sept. 14. The Court again informed Plaintiff that he must respond on or before Sept. 30 or face dismissal. Although the notice pleading rules are & We concede that ordinary pleading rules are \textbf{not meant to impose a great burden upon a plaintiff.}\\
\hline
\textbf{Destination Court:} 5th Circuit \newline
\textbf{Destination Date:} 1971-10-21 \newline
\textbf{Source Court:} Supreme Court \newline
\textbf{Source Date:} 1966-06-20 & That petitioners seek to commence an immediate appeal of that portion of the courts order entered on May 28, 1971. The motives of the officers bringing the charges may be corrupt, but that does not show that the state trial court will find the defendant guilty if he is innocent, or that in any other manner the defendant will be & Against any person who is \textbf{denied or cannot enforce in the courts} of such State a right under any law providing for the equal civil rights of citizens of the United States, or of all persons within the jurisdiction thereof;“(2) For any act under color of authority derived from any law providing for equal rights, or for refusing to do any act on the ground that it would be inconsistent with such law. \\
\hline
\textbf{Destination Court:} 9th Circuit \newline
\textbf{Destination Date:} 1980-03-28 \newline
\textbf{Source Court:} Supreme Court \newline
\textbf{Source Date:} 1911-02-20 & In this case there is even a stronger possibility of recurrence since the police have not offered to discontinue the practice. Id. at 43, 65 S.Ct. at 14-15. (Citations omitted). Some might read De Funis v. Odegaard, 416 U.S. 312, 94 S.Ct. 1704, 40 L.Ed.2d 164 (1974), the equal protection challenge to the University of Washington’s “quota” system in admissions as authority for the proposition that the W. T. Grant or the & The questions involved in the orders of the Interstate Commerce Commission are usually continuing (as are manifestly those in the case at bar) and their consideration ought not to be, as they might be, defeated, by short term orders, \textbf{capable of repetition, yet evading review}, and at one time the Government and at another time the carriers have their rights determined by the Commission without a chance of rédress. \\
\hline
\textbf{Destination Court:} 11th Circuit \newline
\textbf{Destination Date:} 2000-03-08 \newline
\textbf{Source Court:} 10th Circuit \newline
\textbf{Source Date:} 1994-11-22 & Section 1512, however, applies to attempts to prevent or influence testimony not only in federal courts but also before Congress, federal agencies, and insurance regulators. Moreover, § 1512(b) subsumes but is significantly broader than the provision of § 1985(2) making it illegal to & Section 1985(2) creates a cause of action against those who \textbf{“conspire to deter, by force, intimidation, or threat}, any party or witness” from attending or testifying in a federal court. \\
\hline
\end{tabular}
\caption{Sample from LePaRD. For readability, only the last few sentence of preceding context are displayed. The portion of the target passage that appears in quotations in the destination opinion is in bold.}
\label{tab:LePaRD_sample}
\end{table*}

% \begin{table*}[b!]
%     \centering %\small{
%     \begin{tabular}{l c | c c | c c }
%         Approach & Number of labels &  \multicolumn{2}{c|}{Development Set} & \multicolumn{2}{c}{Test Set} \\ 
%         & & NDCG@10 & MAP & NDCG@10 & MAP \\ \hline
%         \multirow{3}{*}{BM25} & 10K & 11.38 &  8.85 &  11.37 &  8.88  \\
%         & 20K & 9.53 &  7.40 &  9.56 &  7.39 \\ 
%         & 50K &  7.66 &  5.96 &  7.79 &  6.04 \\ \hline
%         \multirow{3}{*}{SBERT} & 10K &  9.79 &  7.34 &  9.75 &  7.35 \\
%         & 20K &  7.92 &  6.03 &  7.85 &  5.92 \\
%         & 50K &  5.91 &  4.45 &  5.88 &  4.42 \\ \hline
%         \multirow{3}{*}{fine-tuned SBERT} & 10K &  26.31 &  19.88 &  26.27 &  19.84 \\
%          & 20K &  20.8 &  15.47 &  20.72 &  15.32 \\
%          & 50K &  13.98 &  10.31 &  13.78 &  10.12 \\ \hline
%          \multirow{3}{*}{Classification LEGAL-BERT} & 10K & 30.66 & 25.42 & 30.75 & 25.52 \\
%          & 20K & 23.49 & 19.43 & 23.77 & 19.71 \\
%          & 50K & 15.92 & 13.12 & 16.25 & 13.48 \\ \hline        
%         \multirow{3}{*}{Classification DistilBERT} & 10K &  37.26 &  30.73 &  37.73 &  31.09 \\
%          & 20K &  32.45 &  26.46 &  32.91 &  26.94 \\
%          & 50K &  24.21 &  19.69 &  24.57 &  20.0 \\ \hline        
%     \end{tabular} %}
%     \caption{Additional results: NDCG@10 and Mean Average Precision for development and test set using various baselines. Best results were obtained using classification and (relatively) few labels. Metrics calculated using the \cite{VanGysel2018pytreceval} package.}
%     \label{tab:additional_results}
% \end{table*}

\end{document}